\PassOptionsToPackage{numbers,compress}{natbib} % Force numeric citations

\documentclass{article}

\usepackage{wrapfig}

\usepackage[final]{neurips_2024}
\PassOptionsToPackage{numbers,compress}{natbib}

\usepackage{graphicx}

\usepackage[utf8]{inputenc} 
\usepackage[T1]{fontenc}
\usepackage{hyperref}
\usepackage{url}
\usepackage{booktabs}
\usepackage{amsfonts}
\usepackage{nicefrac}
\usepackage{microtype}
\usepackage{xcolor}
\usepackage{wrapfig}

\usepackage[utf8]{inputenc} % allow utf-8 input
\usepackage[T1]{fontenc}    % use 8-bit T1 fonts
\usepackage{hyperref}       % hyperlinks
\usepackage{url}            % simple URL typesetting
\usepackage{booktabs}       % professional-quality tables
\usepackage{amsfonts}       % blackboard math symbols
\usepackage{nicefrac}       % compact symbols for 1/2, etc.
\usepackage{microtype}      % microtypography
\usepackage{xcolor}         % colors

\title{\Large Removing neural signal artifacts with autoencoder-targeted adversarial transformers (AT-AT)}

\author{%
  Benjamin J. Choi\\
  Kempner Institute at Harvard University\\
  150 Western Ave, Boston, MA 02134\\
  \texttt{benchoi@college.harvard.edu} \\
}

\begin{document}

\maketitle

\begin{abstract}

Electromyogenic (EMG) noise is a major contamination source in EEG data that can impede accurate analysis of brain-specific neural activity. Recent literature on EMG artifact removal has moved beyond traditional linear algorithms in favor of machine learning-based systems. However, existing deep learning-based filtration methods often have large compute footprints and prohibitively long training times. In this study, we present a new machine learning-based system for filtering EMG interference from EEG data using an autoencoder-targeted adversarial transformer (AT-AT). By leveraging the lightweight expressivity of an autoencoder to determine optimal time-series transformer application sites, our AT-AT architecture achieves a >90\% model size reduction compared to published artifact removal models. The addition of adversarial training ensures that filtered signals adhere to the fundamental characteristics of EEG data. We trained AT-AT using published neural data from 67 subjects and found that the system was able to achieve comparable test performance to larger models; AT-AT posted a mean reconstructive correlation coefficient above 0.95 at an initial signal-to-noise ratio (SNR) of 2 dB and 0.70 at -7 dB SNR. Further research generalizing these results to broader sample sizes beyond these isolated test cases will be crucial; while outside the scope of this study, we also include results from a real-world deployment of AT-AT in the \emph{Appendix}. 

\end{abstract}

\section{Introduction}

Advanced brain-computer interfaces (BCIs) promise to unlock exciting possibilities in both rehabilitation \citep{Huang2022} and augmentation \citep{Cinel2019}. Specific potential applications include neuroprosthetic devices \citep{Choi2025a}, nonverbal communication systems \citep{Kashihara2014}, paralysis aids \citep{Chaudhary2021}, and beyond. However, poor signal quality—particularly for noninvasive brain-computer interfaces that attempt to avoid intracranial implantation processes \citep{Sun2018}—presents a major barrier to deployment. In this paper, we seek to improve the signal quality of non-invasive techniques like electroencephalography (EEG) by eliminating the primary threat of electromyographic (EMG) interference \citep{Liang2021}. A more comprehensive examination of this issue, alongside end-use applications and geometric machine learning methodologies, is provided in \cite{Choi2025b}. 

Conventional algorithmic solutions for filtering EMG interference in raw EEG largely center upon linear signal processing methods like canonical correlation analysis (CCA) and independent component analysis (ICA) \citep{Krim1996, iriarte2003, DeClercq2006}, which can be inadequate in elaborate single-channel source separation scenarios \citep{Zhang2021}. The underlying hypothesis guiding much of the work presented herein—in the lineage of prior work \citep{Choi2025c}—is that ML-based signal processing algorithms possess the requisite expressivity to potentially handle complex single-channel interference cases. Yet while recent progress in the field has demonstrated promise in terms of ML-based signal processing \citep{Zhang2021ICASSP, Yin2023, Cui2024}, issues remain in terms of expensive model footprints, latency issues, and undesirable levels of signal degradation. As emphasized in prior art \citep{Wang2020, Pacini2024}, large model footprints can be prohibitive for many real-world, resource-constrained BCI applications.

While CNN-based EEG-EMG signal filtration algorithms take advantage of the spatial locality inherent in EEG processing \citep{Zhang2021ICASSP}, transformers can harness both the spatial locality and non-locality relevant to neural signal reconstruction \citep{Vaswani2017, Cho2023}. Prior literature exists on the application of transformers to EEG data \citep{Pu2022, Wang2024}; on the whole, however, transformer denoising results remain suboptimal compared to the state-of-the-art (SOTA) CNN-based approach \citep{Zhang2021ICASSP}. We anticipate that the development of more effective transformer application site determination methods—as well as non brute-force approaches to tokenization and target site optimization—may enable superior downstream performance; these ideas have yet to be rigorously considered in EEG transformer literature. 

In order to rectify both the (1) sub-ideal reconstructive accuracy of existing transformer-based neural signal filters (especially resulting from selective application and residual reconstructive aberrations), and (2) inefficient tokenization and target site optimization methods, we thus present the following:
\begin{enumerate}
\item To address point 1, we present the marriage of a signal-processing transformer with the principles of a generative adversarial network \citep{Creswell2018}. By training the generative transformer in an adversarial environment, the network output can be more carefully regulated—and thus made to precisely mimic that of pristine neural signal (leading to superior reconstructive accuracy). While adversarial transformers have been demonstrated in disjoint generative applications \citep{Hudson2021, Kasem2019}, they have yet to be applied to neural blind source separation. 
\item To address point 2, we present a target site determination method using a lightweight denoising autoencoder as a noise assessment proxy. The determined “noise level” will be used to determine transformer application sites and tokenize the input stream; high-noise sites will be masked and the time series transformer will be subsequently invoked to reconstruct areas of interest.
\end{enumerate}

\begin{figure}[h]
  \centering
  \includegraphics[width=1\columnwidth]{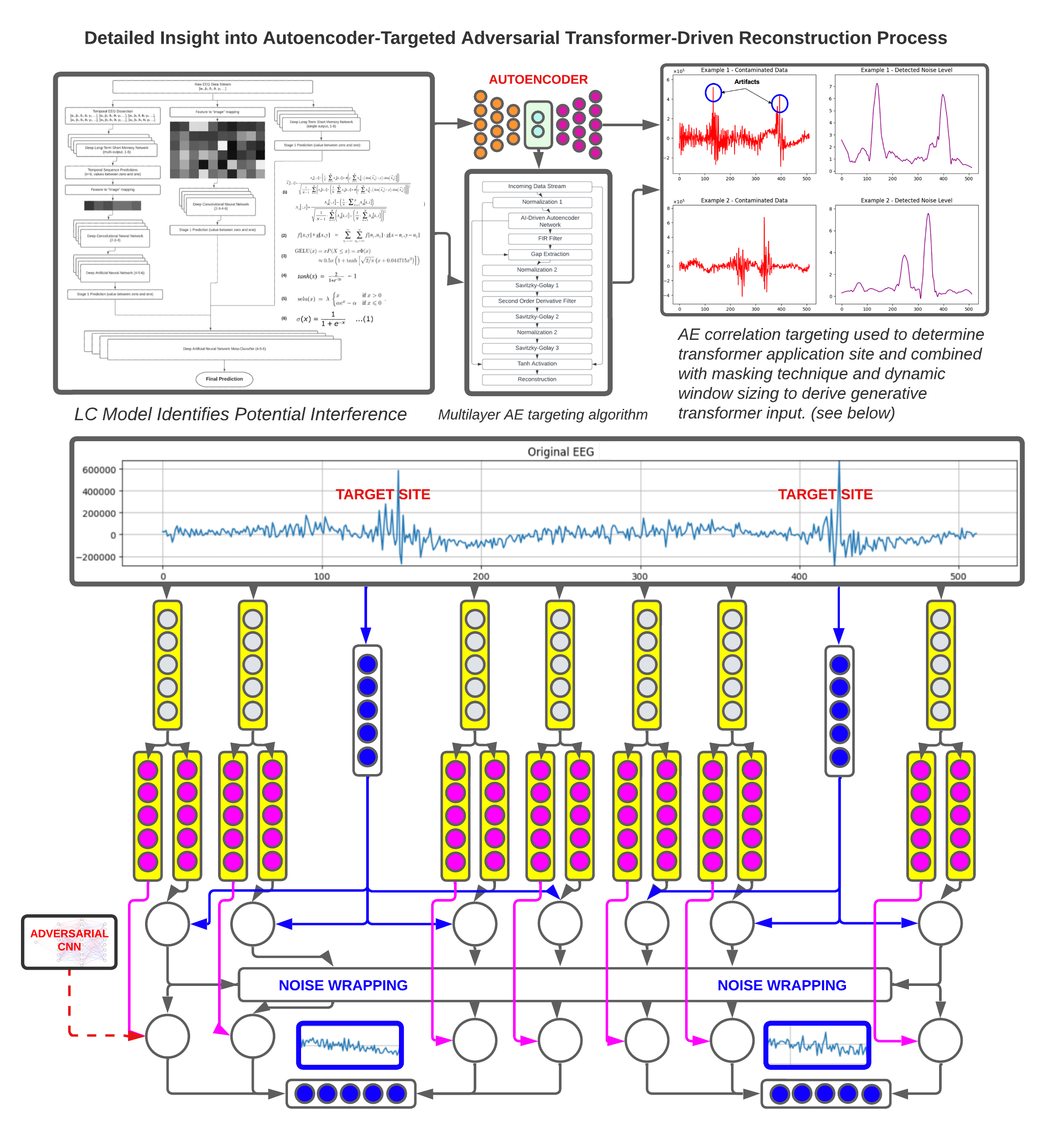}
  \caption{The autoencoder-targeted adversarial transformer (AT-AT) system architecture. The initial signal is passed to the LSTM-CNN (LC) targeting model and denoising autoencoder (upper panels); the transformer is subsequently invoked on select target sites (lower diagram). Design incorporates elements featured in \cite{Choi2025b, Choi2025c}.}
  \label{fig:at-at-results2}
\end{figure}

The subsequent \emph{autoencoder-targeted adversarial transformer system (AT-AT)} will be employed—as a lightweight, autoencoder-connected solution—to remediate the primary noise barrier inhibiting effective neural interfaces. In this study, we build an AT-AT model and validate AT-AT's denoising performance in a standardized semi-synthetic setting on a major benchmark blind source separation (BSS) dataset. Though outside the primary scope of this study, the \emph{Appendix} also provides a preliminary validation of AT-AT performance in a real-world setting (with no ``ground truth" signal present); this real-world application (in conjunction with a manifold learning pipeline) is formally covered in depth in \cite{Choi2025b}.

\section{Methods}

In order to validate the proposed AT-AT denoising system, we employed the benchmark EEGdenoiseNet dataset \citep{Zhang2021}. While other available datasets were surveyed \citep{Koelstra2011, VanVeen2019, Kaya2018}, EEGdenoiseNet has emerged in the BCI field as a major benchmark for training and validating published ML denoising algorithms \citep{Zhang2021, Zhang2021ICASSP, Yin2023, Cui2024}, thereby enabling objective comparison of AT-AT against current gold standards. Moreover, EEGdenoiseNet surpasses alternatives in terms of both overall scope and the ability for ground truth verification—crucial for benchmark testing. We conduct a preliminary validation of AT-AT in line with prior autoencoder-based work \citep{Choi2025c}, using delineated methods from \cite{Zhang2021} to construct a semi-synthetic contamination problem. We built and trained the AT-AT system (\emph{Figure 1}) on two train mixtures and two test mixtures at -7 dB and 2 dB with an EEG-EMG BSS model objective. As alluded to in the \emph{Introduction}, a major engineering goal of the AT-AT system was to demonstrate a targeted application of attention networks for filtration with a constrained model footprint and training time. However, to appropriately assess model training time in a live BSS context (as previously noted in \cite{Choi2025c}), one must consider not only model training time but also the total combined duration of EEG and EMG segments used in a single training run. In other words, the model training time is not merely the training length itself, but rather the maximum of the training length \emph{and} the duration of neural data needed for a complete training run. In light of this, for the AT-AT system, we restricted training data to 120 EEG segments contaminated with high-variance EMG artifacts totaling 240s in combined duration. Thus, four minutes' worth of data are required to support the AT-AT training runs described in this study. 

In accordance with methods for inferring signal SNR before running a subsequent tailored denoising model, we first run an initial upstream model based on a hybrid LSTM-CNN (LC) architecture to determine the appropriate SNR target level \citep{Choi2025c}; details of this upstream model are included in the \emph{Appendix}. Subsequently, as a first denoising pass, the AT-AT architecture employs a convolutional autoencoder with batch normalization; this architecture showed promise for EEG-EMG denoising in prior work \citep{Choi2025c}. This autoencoder structure performs reasonably well in lower-noise settings but still demonstrates suboptimal performance on especially noisy reconstruction sites. Thus, to rectify the shortcomings of this convolutional autoencoder, we selectively invoke a time-series transformer model to handle more intensive instances of reconstruction. Introducing a second model architecture comes at the cost of a more complex parameter space search and longer training run, so the time-series transformer is only used when necessary in order to accommodate a more intensive retraining process. This novel hybrid use of autoencoders and transformers in conjunction is a key component of the AT-AT system; an evaluation of a sans-transformer method is outside of the scope of this study, but we refer readers to \cite{Choi2025c}.

However, the notion of selective invocation of the time-series transformer begs a key question: \emph{when} (and based on what criteria) should the transformer be invoked? This paper introduces a new criterion for transformer invocation based on a heuristic for assessing the noise present in an EEG signal. Specifically, we employ the \emph{correlation between the denoised autoencoder output and the original signal} as a proxy to assess estimated noise. Given that the autoencoder is trained to output ``pristine" EEG, we infer that if the autoencoder outputs a stream highly correlated with the original raw signal, the original signal itself contains relatively less noise; conversely, if the autoencoder output differs (in terms of correlation coefficient) from the original signal, then we infer relatively more noise to be present. These ``high-noise" sites determined by the \emph{autoencoder-targeting} heuristic are then masked and ``filled in" by the time-series transformer model. In support of this hybrid autoencoder-transformer approach, careful pains were taken to set up the details of proper autoencoder-based tokenization and masking. Time--series tokens were formed by combining the digital sampling values from the original and autoencoder datastreams into 1x2 tokens encapsulating the joint information. Token values corresponding to high-noise sites were then masked according to the aforementioned methodology, with a 0.8 correlation coefficient (CC) between original and autoencoder-filtered signal set as a cutoff delineating low-noise and high-noise sites.

\begin{figure}[h]
  \centering
  \includegraphics[width=1\columnwidth]{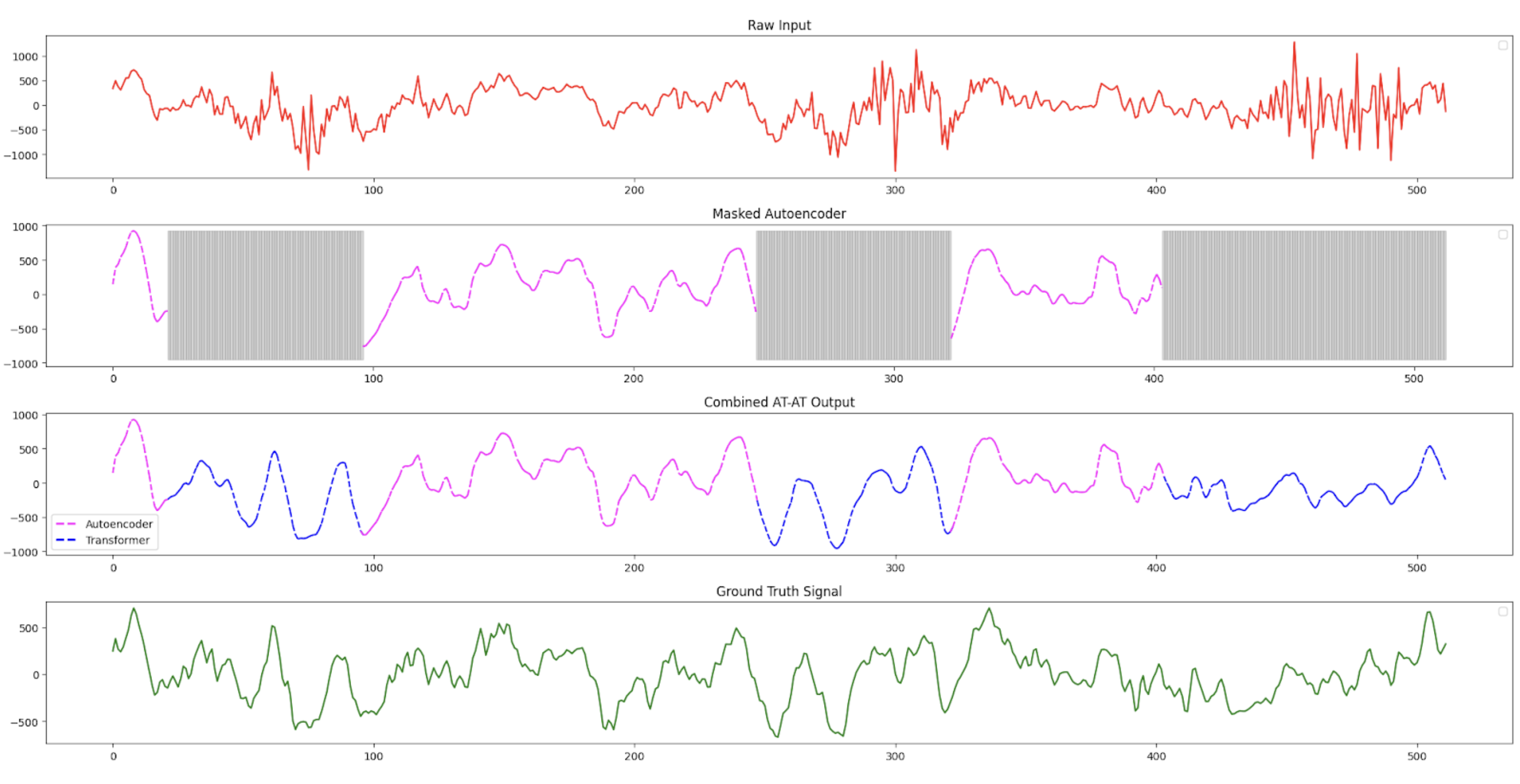}
  \caption{AT-AT model workflow. From top to bottom: (1) raw input signal, (2) initial autoencoder filtration pass with high-noise target site masking, (3) time-series transformer-based reconstruction of target sites, (4) ground truth signal.}
  \label{fig:at-at-results9}
\end{figure}

One key challenge stemming from this combination of autoencoder and transformer architectures was that of spectral preservation. The selective insertion of time-series transformer-generated signals can introduce irregularities into the model output and prevent the model output characteristics from appropriately resembling the spectral characteristics of authentic EEG data. To achieve a well-regulated output compliant with the physical properties of EEG, we implemented an adversarial training component to our autoencoder-transformer hybrid architecture. Specifically, as mentioned in the \emph{Introduction}, we treated the transformer-augmented autoencoder output as the ``generator" and implemented a convolutional neural network-based (1D-CNN-based) discriminator model to distinguish between authentic versus model-reconstructed EEG. This generator-discriminator pair was folded into a five-cycle GAN training loop; ablation-wise, in line with prior work \citep{Choi2025c}, we found that the addition of adversarial training led to $>$10\% gains in spectral filtration accuracy.

After building and validating the AT-AT system across a partitioned 100-sample test set on the benchmark semi-synthetic contamination problem, we also implemented additional end-use case validation. In the EEGdenoiseNet context, performance can be assessed via direct comparison with the synthetically contaminated ground truth samples. But in real-world environments, the only way to assess the performance of our signal processing pipeline is by measuring improvement on downstream classification tasks. Thus, in order to demonstrate the potential of an AT-AT model at unlocking better BCI outcomes, we deployed AT-AT onto a notable BCI use case: digit- versus non-digit classification on the \emph{MindBigData} EEG dataset \citep{Vivancos2022} (see \emph{Appendix} for further details). This downstream application ultimately incorporated an extensive geometric machine learning pipeline which is discussed in \cite{Choi2025b} (but falls outside of the scope of this paper).

\section{Results \& Discussion}

AT-AT posted a mean reconstructive correlation coefficient (CC) with ground truth of 0.951 at 2 dB (95\% CI: 0.947, 0.954), a temporal relative root mean square error (tRRMSE) of 0.317 (95\% CI: 0.305, 0.329), and a spectral relative root mean square error (sRRMSE) of 0.270 (95\% CI: 0.238, 0.303). On the low end of the SNR spectrum (-7 dB), AT-AT posted a CC with ground truth of 0.703 (95\% CI: 0.679-0.726), tRRMSE of 0.759 (95\% CI: 0.731, 0.786), and sRRMSE of 0.800 (95\% CI: 0.753, 0.848). Total mean training time on a T4 High-RAM GPU was measured at 249.1 seconds. \emph{Figure 3} compares AT-AT model performance with existing published deep learning models.

\begin{figure}[h]
  \centering
  \includegraphics[width=0.9\columnwidth]{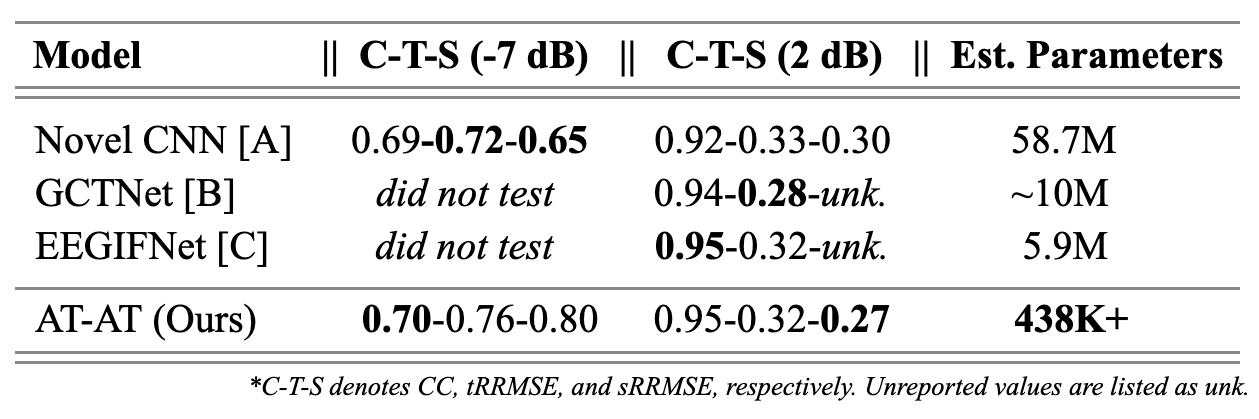}
  \caption{AT-AT performance relative to major deep learning benchmarks. [A], [B], and [C] refer to \cite{Zhang2021ICASSP, Yin2023} and \cite{Cui2024}, respectively. All values are best extrapolated from reported documentation (best-in-class values in bold).}
  \label{fig:at-at-results9}
\end{figure}

While model training times are hardware-dependent, a lower parameter count does imply some conferred benefit in the realm of training time and model footprint; as previously mentioned, large parameter counts and model footprints are a limitation of existing high-performance deep learning filtration algorithms. While AT-AT's performance is not dominant across both ends of the SNR spectrum, the model's relatively competitive reconstruction accuracy coupled with a $>$90\% reduction in model size indicate potential benefit in specific use cases where reasonably high BSS performance and low model footprint are important objectives. We also note that the training run time can be regarded as the limiting factor when it comes to dynamic AT-AT training; the model's 249.1 seconds to retrain exceeds its 240-second EEG training data requirement. 

Examining our system on real-world use cases is an additional key component of AT-AT validation. As discussed in the \emph{Appendix} and formally treated in \cite{Choi2025b}, we implemented AT-AT in conjunction with an extensive downstream manifold learning pipeline on the real-world imagined digit- versus non-digit classification problem. The results of this use case deployment (which are informally included in the \emph{Appendix}) offer early evidence that AT-AT may confer utility in real-world BCI settings. 

While promising, it is important to note that all results are preliminary. Crucially, these results will need to be more extensively generalized beyond the initial deployment and semi-synthetic test cases presented in this study. Further training and examination on broader sample sizes and more diverse real-world use cases will be important to assess overall viability.

\section*{Acknowledgements}

This research was accepted for presentation at CNS 2025. The underlying work was generously supported by Prof. Demba Ba and the Kempner Institute at Harvard University via the inaugural KURE research program. 

\bibliographystyle{plain}

\bibliography{references}  

\section{Appendix}

\subsection{Additional Model Specifics}

As previously mentioned, the autoencoder architecture follows a convolutional structure with batch normalization and dropout layers to stabilize training and prevent overfitting. Training is conducted via a correlation coefficient-based objective function. The encoder starts with an input layer that accepts a 1D signal consisting of 512 time steps with a single feature, representing the EEG signal. The encoder includes two convolutional layers: the first Conv1D layer contains 32 filters with a kernel size of 3, followed by batch normalization and ReLU activation. The second Conv1D layer has 64 filters with similar activation and normalization steps. Max pooling with a factor of 2 is applied after each convolution to reduce the temporal dimension. The central autoencoder layer has 128 filters, maintaining the same structure of batch normalization and ReLU activation. On the decoder side, the architecture mirrors the encoder, with upsampling layers replacing the pooling layers. Two convolutional layers reconstruct the signal, and a final Conv1D layer with a single filter and sigmoid activation outputs the final reconstructed signal. The autoencoder is trained using the Adam optimizer with a learning rate of $1 \times 10^{-4}$, a batch size of 20, and 10 epochs of training.

The adversarial model uses the previously described generative adversarial network (GAN) architecture, combining a transformer-based generator and a convolutional discriminator to perform EEG denoising. The generator utilizes a transformer encoder with two encoder layers, multi-head attention (with four heads), and a feedforward network of size 128. Input features, which consist of two channels, are embedded into a 16-dimensional space before entering the transformer. The output from the transformer passes through a convolutional smoothing layer (Conv1D with a kernel size of 3) to ensure smoothness in the generated signal. Finally, a fully connected layer transforms the output into a 1D signal. The discriminator is a Conv1D-based model designed to distinguish between real and generated EEG signals. It contains two convolutional layers: the first Conv1D layer has 64 filters and LeakyReLU activation, followed by dropout, while the second layer has 128 filters and the same activation and dropout sequence. A fully connected layer reduces the representation to a scalar, which is processed by a sigmoid activation function to classify the signal as authentic or generated. The GAN training process alternates between generator and discriminator updates over five cycles in each iteration, with both models using binary cross-entropy loss for optimization. The Adam optimizer is used for both the generator and discriminator, with a learning rate of $1 \times 10^{-4}$. The adversarial model is trained over 10 epochs with a batch size of 20.

As mentioned in the \emph{Methods} section, we trained an upstream pre-processing LC model to infer the appropriate SNR target level before deploying AT-AT, following previously demonstrated methods \citep{Choi2025c}. This model selects the suitable iteration of AT-AT based on the detected SNR level, enabling more tailored processing. Pre-processing accounted for 32.9\% of the 249-second training time. The model integrates a hybrid architecture combining LSTM and CNN pathways that feed into a meta-classifier. The CNN pathway reshapes the EEG input into 2D blocks and processes them through Conv2D layers with ReLU, batch normalization, max-pooling, and dropout. The LSTM pathway captures temporal dependencies from the raw EEG data through two LSTM layers, which are flattened. The LSTM-CNN-MLP pathway involves two LSTM layers reshaped for convolutional processing via Conv2D followed by a dense layer. These outputs are concatenated into a meta-classifier with two fully connected layers and a softmax output, which predicts the appropriate SNR level across the -7 to 2 dB range. The training process spans 100 epochs with a batch size of 100. Inferring relative SNR interference has been shown to be a relatively trivial task, with past accuracies demonstrated at 98\% \citep{Soroush2022}; our upstream LC model was able to correctly infer SNR across all 100 test cases. (This pre-processing layer is used to toggle between downstream models and does not directly interact with the reconstruction process.)

\subsection{Post-Processing for BCI Deployment on the Imagined Digit Classification Problem}

As previously discussed, in addition to evaluating denoising performance on the semi-synthetic EEGdenoiseNet problem, we conducted a preliminary assessment of AT-AT's efficacy on the real-world MindBigData \citep{Vivancos2022} use case. This use case primarily consisted of the classification of two-second raw EEG segments corresponding to digit- versus non-digit-related thoughts into their appropriate categories. Methodologically, we partitioned the MindBigData classification problem via a 90:10 train-test split and ran the entire raw neural dataset through the AT-AT architecture. The resulting improvement in class separability achieved solely via AT-AT (with an unmodified default t-SNE configuration) is depicted in \emph{Figure 4}. After denoising the neural signals via AT-AT, we developed an extensive geometric machine learning pipeline, which is detailed in \cite{Choi2025b}.

\begin{figure}[h]
  \centering
  \includegraphics[width=1\columnwidth]{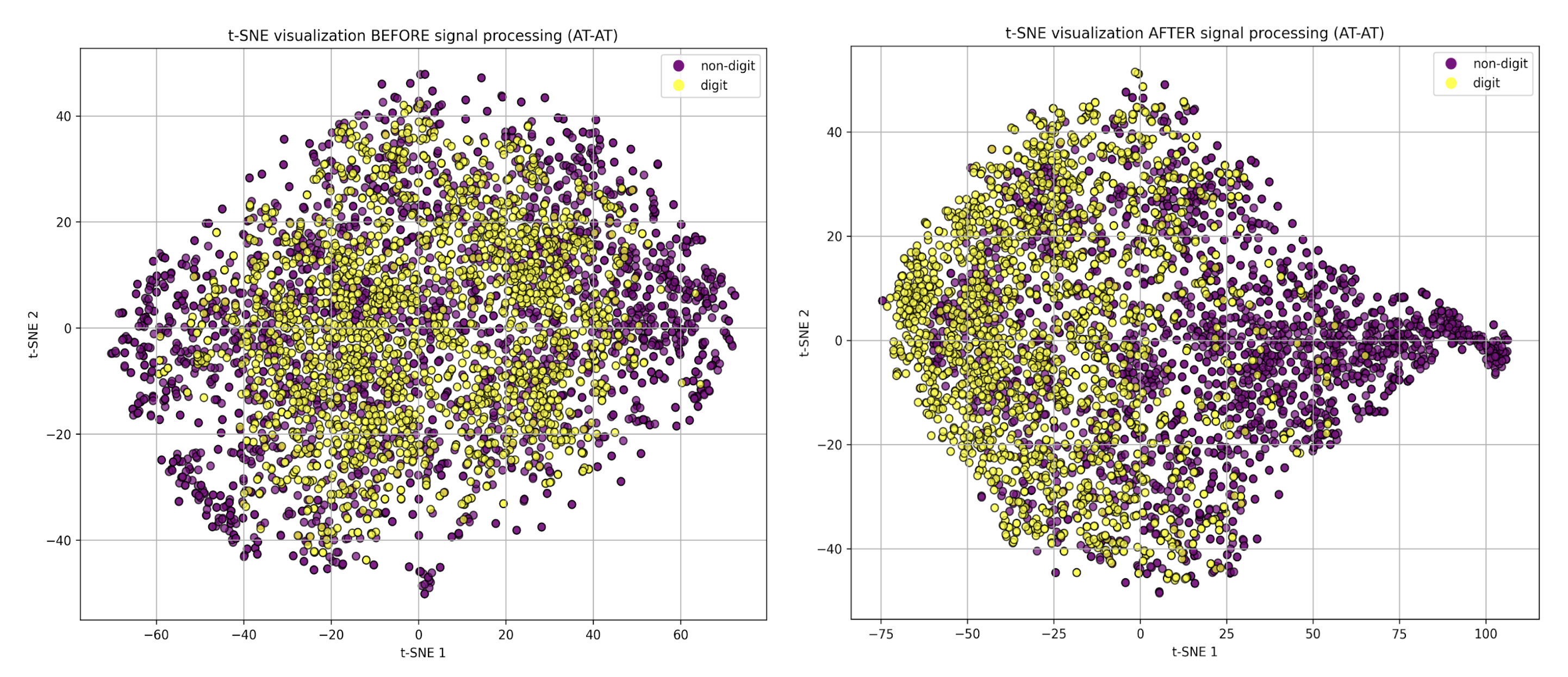}
  \caption{AT-AT processing enables separability between neural signal classes (i.e., digit- vs. non-digit-related thoughts) to be elucidated. Pre-AT-AT t-SNE is depicted left, while the post-AT-AT t-SNE is depicted right; also given in \cite{Choi2025b}.}
  \label{fig:at-at-results2}
\end{figure}

In short, after filtering the neural signals through the AT-AT architecture, we first ran each channel through a one-sided FFT to generate frequency vectors and performed dimensionality reduction using a Laplacian eigenmap. This combined FFT and Laplacian eigenmap representation was further refined using a discrete Ricci flow analog (via Ollivier-Ricci curvature) in accordance with methods described in \cite{Weber2017} and \cite{Tian2023} to capture nuanced spatial relationships between EEG channels. After deriving this optimal graph structure, we then trained a graph convolutional network (GCN) with skip connections over 100 epochs to preserve graph distances and perform dimensionality reduction. A lightweight, 5.2K-parameter 1D-CNN classifier, trained over 70 epochs, was then used for final classification on the GCN-reduced data. 

Overall, we observed significant improvement on the order of a $>$40\% reduction in classification error after AT-AT filtration (in line with the increased class separability shown in \emph{Figure 4}) on EEG-based classification of ``digit" versus ``non-digit" thoughts. When combined with downstream manifold learning and post-processing, our AT-AT processing layer enabled digit- versus non-digit-related thoughts to be classified with 97.0\% test accuracy (95\% CI: 93.66, 100.0). These results offer preliminary evidence that AT-AT, beyond its performance on the benchmark semi-synthetic source separation task, confers some utility in real-world BCI settings. 

We also note that the t-SNE plots observed in \emph{Figure 4} may offer a preliminary suggestion that the underlying \emph{semantic meaning vectors} corresponding to user thoughts could be genuinely manifested in the geometric structure—as elucidated via AT-AT processing—of neural signal patterns. This implication could open the door to further studies on the geometric modeling of semantic relationships in EEG brainwave data (as potentially unraveled by autoencoder-targeted adversarial transformer-based models). 

\end{document}